\documentclass[sigconf,nonacm]{acmart}

\renewcommand\footnotetextcopyrightpermission[1]{}

\usepackage{booktabs}
\usepackage{xcolor}
\usepackage{pifont}
\usepackage{makecell}
\usepackage{multirow}
\usepackage{algorithm}
\usepackage{algpseudocode}

\definecolor{mygreen}{RGB}{0,150,0}
\newcommand{\cmark}{\textcolor{mygreen}{\ding{51}}}
\newcommand{\xmark}{\ding{55}}

\algrenewcommand\algorithmicrequire{\textbf{Input:}}

\newcommand{\equalcontrib}{\textsuperscript{*}}

\begin{document}

\title{CLDRoute: Conditional Latent Diffusion for Routability Map Generation in Physical Design}


\author{%
  Kiran Thorat\textsuperscript{1}\equalcontrib
  \quad
  Nicole Meng\textsuperscript{2}\equalcontrib
  \quad
  Caiwen Ding\textsuperscript{3}
  \quad
  Yingjie Lao\textsuperscript{2}
  \quad
  Zhijie Jerry Shi\textsuperscript{1}
}
\vspace{2mm}

\affiliation{%
  \institution{%
  \vspace{2mm}
    \textsuperscript{1}University of Connecticut
    \qquad
    \textsuperscript{2}Tufts University
    \qquad
    \textsuperscript{3}University of Minnesota Twin Cities
    \\[0.45em]
    \texttt{\{kiran\_gautam.thorat,zshi\}@uconn.edu}
    \quad
    \texttt{\{ziyi.meng,Yingjie.lao\}@tufts.edu}
    \quad
    \texttt{dingc@umn.edu}
  }
  \city{}
  \country{}
}

\begin{abstract}

Accurate routability estimation during physical design is important for reducing costly post-routing iterations. Prior learning-based methods treat this task as deterministic prediction, mapping placement-stage features to a single congestion or DRC outcome. We instead formulate routability estimation as a \emph{conditional generation} problem, where both routing congestion and DRC violations are modeled as spatially structured \emph{routability fields}.
Our framework, \emph{\underline{C}}onditional Latent Diffusion for \emph{\underline{Route}}ability estimation (\emph{CLDRoute}), uses physics-aware conditioning and task-specific latent modeling to handle the different characteristics of congestion and DRC maps.
This allows our method to supports sample-based inference, producing both a mean prediction and a spatial uncertainty estimate for the same input design. On CircuitNet~2.0 (N28), our method achieves, for DRC violation generation, an SSIM of 0.9678, an MAE of 0.0028, and a TopK@1\% of 0.3494; for congestion generation, it achieves an SSIM of 0.9031, an MAE of 0.0286, and an NZ-Pearson of 0.3692. Overall, our framework provides a more practical view of routability at placement by generating both the expected outcome and its uncertainty.

\end{abstract}


\maketitle

\begingroup
  \renewcommand{\thefootnote}{*}
  \footnotetext[0]{Equal Contribution.}
\endgroup
\footnotetext[1]{%
  Code is available at
  \href{https://github.com/kiranthorat3/CLDRoute}
       {\nolinkurl{github.com/kiranthorat3/CLDRoute}}.
}

\section{Introduction}
\label{sec:intro}

Routability estimation at the placement stage has received increasing attention, as routing congestion and design-rule-check (DRC) violations identified after routing often trigger costly iterations across placement, routing, and verification. As design rules and interconnect complexity continue to increase, late-stage failures remain a major source of design closure cost. This has motivated substantial research on predicting routing outcomes prior to detailed routing.
~\cite{chai2023circuitnet,jiang2024circuitnet}.

Existing learning-based methods primarily formulate routability as a prediction task. Given a design representation, the model predicts a routing congestion map, a DRC-related target, or a post-routing property~\cite{8587655,zou2025lay}. However, the dominant formulation remains to regress a single output map or scalar target from a placed design. 

The main gap lies in the formulation itself. At the placement stage, the final routing solution is not yet available. A placed design can admit multiple valid downstream routing outcomes, as congestion resolution, detours, layer assignment, and detailed routability constraints are only resolved during routing. Routability is therefore not a single fixed target at placement, but rather a conditional distribution over plausible routing outcomes. A deterministic predictor compresses this distribution into a single estimate and cannot capture spatial uncertainty \cite{zhao2025deeplayout, zou2025lay,9473959}. In this sense, routability is better formulated as \emph{conditional generation} rather than deterministic regression. In this work, we model both the routing congestion map and the DRC violation map as generated \emph{routability fields}. 


A second gap lies in the conditioning representation. Many existing methods rely on a limited set of placement-derived channels, such as macro-region and RUDY-style features~\cite{zou2025lay,chai2023circuitnet,jiang2024circuitnet}. In this work, we introduce \emph{physics-aware conditioning channels} to guide the model using routing-relevant spatial signals that are more directly tied to routing demand and available resources. The goal is not only to predict a single map, but to generate a distribution of plausible routing outcomes under physically meaningful conditioning.

Based on these observations, we propose a conditional generative framework for routability estimation. The framework models both the routing congestion map and the DRC violation map within a unified pipeline, while allowing the two targets to retain their distinct output characteristics. Instead of returning a single prediction, it supports sample-based inference, producing both a mean prediction and a spatial uncertainty map. The goal is to move routability estimation from single-output prediction to distributional modeling that better reflects the nature of the physical-design problem.
The main contributions are summarized as follows:
\begin{itemize}

\item \textbf{Distributional formulation of routability:} Routability is designed as conditional generation rather than deterministic prediction in our method, modeling congestion and DRC as stochastic \emph{routability fields}. This formulation captures the inherent ambiguity of placement-stage signals and exposes spatial uncertainty that is inaccessible to regression-based methods.

\item \textbf{Physics-aware spatial conditioning:} CLDRoute's conditioning is structured through routing-relevant signals tied to demand and resource constraints, moving beyond generic placement features. This provides direct control over the generation process and grounds predictions in physically meaningful cues.

\item \textbf{Task-aligned latent modeling:} Congestion and DRC targets are represented in separate latent spaces matched to their distinct statistical properties, allowing high-resolution modeling without sacrificing efficiency or fidelity.

\item \textbf{Controllable generative pipeline with uncertainty:} A multi-scale ControlNet-style latent diffusion framework produces both accurate predictions and calibrated spatial uncertainty maps, offering actionable signals for identifying high-risk regions during design.
\end{itemize} 

As shown in Table~\ref{tab:comparison_methods}, existing methods primarily map placement-stage features to a single deterministic routability output. In contrast, CLDRoute models routability as a conditional distribution over plausible congestion and DRC maps, and produces spatial uncertainty through sample-based inference. To the best of our knowledge, CLDRoute is the first framework to formulate placement-stage routability estimation in this way for both tasks. This formulation better reflects the ambiguity of routing outcomes at placement stage and motivates the physics-aware, task-specific design of our framework

\section{Background }
\label{sec:back}
\begin{table}[t]
\centering
\caption{Comparison of routability methods.}
\label{tab:comparison_methods}
\scriptsize
\setlength{\tabcolsep}{6pt}
\begin{tabular}{lcccc}
\toprule
Method &
\makecell{Physics-aware\\controls} &
\makecell{Both\\tasks} &
\makecell{Conditional\\generation} &
\makecell{Spatial\\uncertainty} \\
\midrule
LayNet~\cite{zou2025lay}    & \xmark & \cmark & \xmark & \xmark \\
GPDL~\cite{9473959}         & \xmark & \cmark & \xmark & \xmark \\
RouteNet~\cite{8587655}     & \xmark & \xmark & \xmark & \xmark \\
CLDRoute (ours)             & \cmark & \cmark & \cmark & \cmark \\
\bottomrule
\end{tabular}
\end{table}

The physical design stage in EDA involves large-scale NP-hard optimization, where placement and routing jointly determine design quality~\cite{Mirhoseini2021AGP}. Routing resolves wire assignments, layer usage, and design-rule constraints, and is often the most time-consuming stage. Predicting routing outcomes at placement has therefore become important for reducing costly design iterations.

\subsection{Routability Estimation}

Routability estimation aims to predict post-routing properties, such as routing congestion and design-rule-check (DRC) violations, from placement-stage representations. These predictions guide placement optimization by identifying regions likely to cause routing failures. A common representation is the routing congestion map, which captures routing demand relative to available resources across the layout grid, while DRC-related targets identify potential violations under detailed routing constraints.

Most existing approaches treat routability estimation as a supervised prediction task, where a model maps placement-derived features to a single congestion map or scalar metric~\cite{xie2018routenet}. However, this formulation assumes a one-to-one mapping between placement and routing outcomes, which does not reflect the variability introduced by routing decisions such as detours, layer assignment, and local constraint resolution.

\subsection{Learning-based Routability Prediction}

\noindent \textbf{DNN-based methods.}
Deep learning approaches typically convert layouts into image representations and apply fully convolutional networks (FCN) to predict routability~\cite{8587655, 9473959}. For example, GPDL~\cite{8587655} adopts an encoder--decoder architecture to map grid features to congestion maps, while RouteNet~\cite{xie2018routenet} uses CNN-based modeling for routing-related prediction. Generative models in computer vision ~\cite{meng2025advancing} such as GANs~\cite{yu2019painting, liang2022stochastic} and conditional graph attention networks~\cite{yu2019painting} have also been explored for congestion forecasting. However, these methods are often limited by resolution and by the use of incomplete domain-specific features.

\noindent \textbf{GNN-based methods.}
Graph neural network approaches construct graphs from netlists and layout features, encoding both connectivity and spatial information~\cite{yang2022versatile, ghose2021generalizable, baek2022pin}. These methods leverage both topological and geometric signals to predict congestion and related metrics. While they provide a more structured representation, their effectiveness depends on graph construction quality and they may not explicitly model routing resource constraints.

Across both image-based and graph-based methods, they focuses mainly on deterministic prediction of a single output. This formulation does not capture the inherent uncertainty of routing outcomes at the placement stage, where multiple valid routing solutions may exist. In addition, conditioning is often treated as an unstructured collection of input channels, without explicitly incorporating routing-relevant physical signals. These limitations motivate a formulation that models routability as a distribution over plausible outcomes, conditioned on physically meaningful design features.

\section{Framework}
\begin{figure*}[t]
    \centering
    \includegraphics[width=0.8\textwidth]{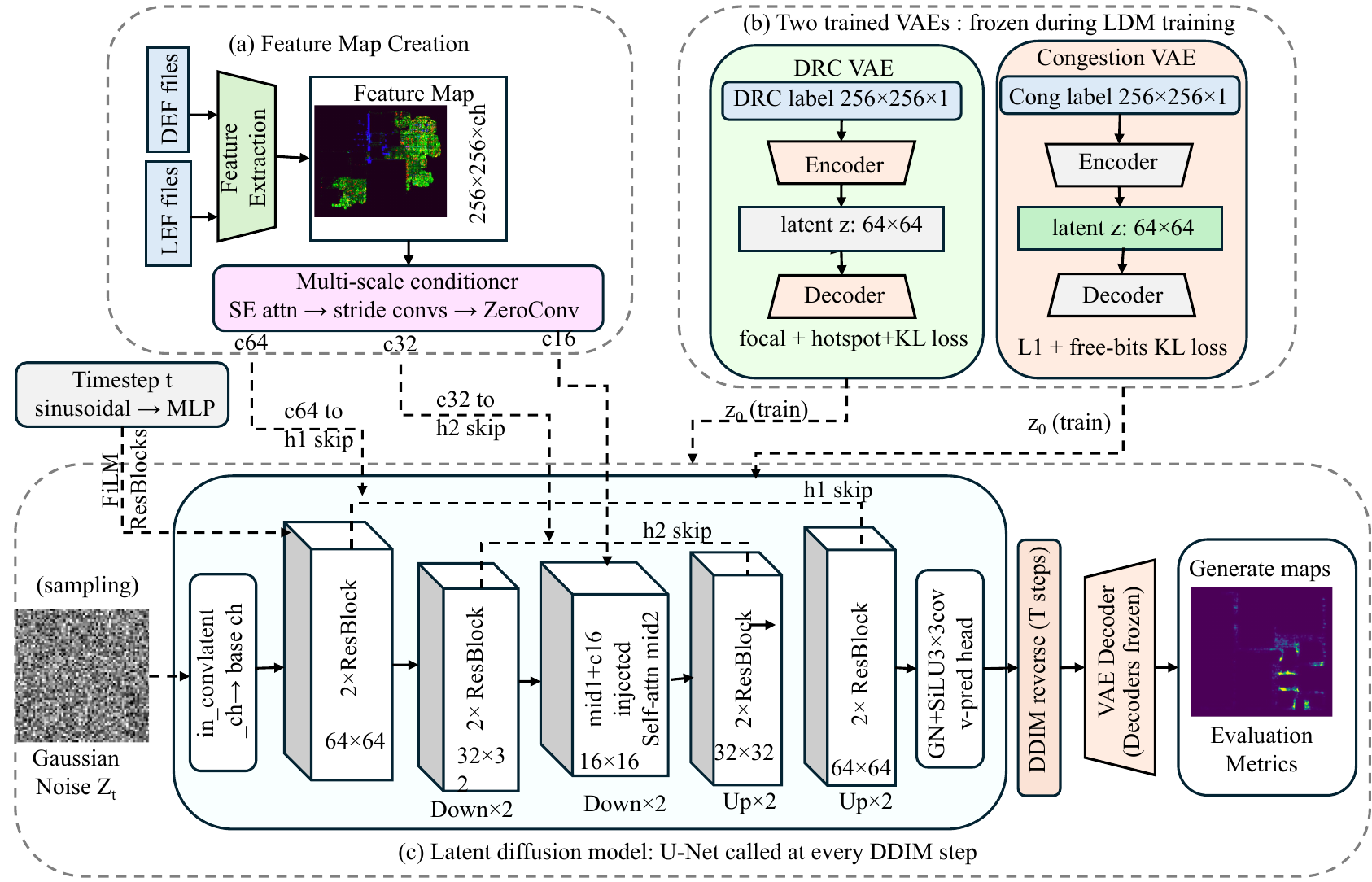}
    \caption{Overview of the proposed conditional latent diffusion framework for routability prediction.
    (a) Routing features extracted from DEF/LEF files are passed through a multi-scale conditioner producing spatial conditioning signals $\mathbf{c}_{64}$, $\mathbf{c}_{32}$, and $\mathbf{c}_{16}$.
    (b) Two task-specific VAEs compress DRC and congestion labels into latent representations and are frozen during LDM training.
    (c) A conditioned U-Net denoiser generates latent codes via DDIM sampling with classifier-free guidance; the frozen VAE decoder produces the final routability map.}
    \label{fig:overview}
\end{figure*}

As illustrated in Fig.~\ref{fig:overview}, the framework proceeds in three stages. Fig.~\ref{fig:overview}(a) shows how placement-stage design files are converted into spatial routing controls. Fig.~\ref{fig:overview}(b) shows the two task-specific VAEs used to encode DRC and congestion maps into separate latent spaces. Fig.~\ref{fig:overview}(c) shows the conditional latent diffusion model, where the noisy latent is denoised under multi-scale routing controls and then decoded to produce the final routability map.

\subsection{Physics-Aware Routing Controls}
\label{sec:features}

\begin{figure*}[t]
    \centering
    \includegraphics[width=.80\linewidth]{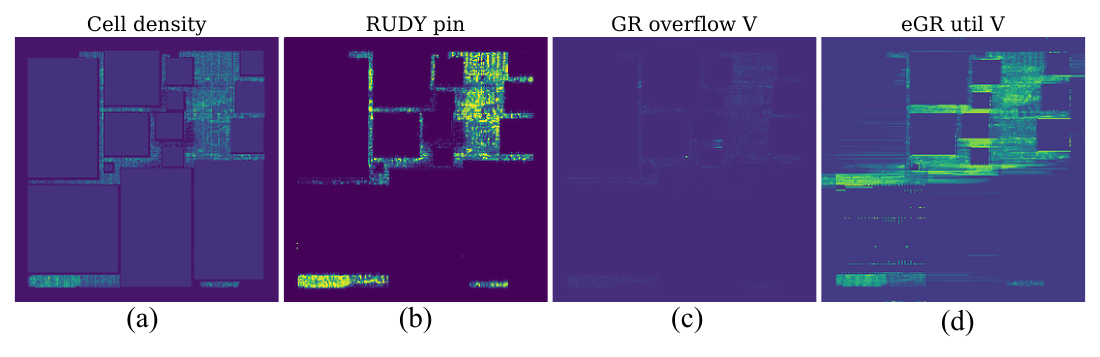}
    \caption{Representative DRC routing controls for one test design: (a) cell density, (b) \texttt{RUDY\_pin}, (c) \texttt{GR\_overflow\_V}, and (d) \texttt{eGR\_util\_V}. Together, these controls provide geometry, demand, and supply information used by the model.}
    \label{fig:drc_controls}
\end{figure*}

The quality of conditional generation depends directly on the quality of its routing controls. Prior routability models usually reuse feature sets from benchmark pipelines, where the channels are selected mainly for availability and for regression-based prediction~\cite{chai2023circuitnet,jiang2024circuitnet,8587655,9473959}. In this work, we instead design the routing controls from physical-design reasoning. The objective is not simply to add more channels, but to expose the model to spatial signals that are directly related to routability.

Congestion and DRC depend on different physical cues, so they should not use the same routing controls. Congestion is governed mainly by the spatial balance between routing demand and available routing resources across the layout. DRC depends more strongly on local access pressure, routing conflicts, and geometric context near obstacles and dense placement regions. This motivates task-specific control design: congestion uses controls that emphasize layout-wide demand--resource balance, while DRC uses a broader set of local conflict cues.

We organize the routing controls into three physical groups,
\begin{equation}
\mathcal{F}_{\mathrm{task}}
=
\mathcal{F}_{\mathrm{demand}}
\cup
\mathcal{F}_{\mathrm{supply}}
\cup
\mathcal{F}_{\mathrm{geometry}},
\label{eq:feature_groups}
\end{equation}
where $\mathcal{F}_{\mathrm{demand}}$ contains channels that describe where routing pressure originates, $\mathcal{F}_{\mathrm{supply}}$ contains channels that describe the available routing resources and their usage state, and $\mathcal{F}_{\mathrm{geometry}}$ contains channels that describe the placement structure that shapes routing difficulty. In our implementation, the demand group includes RUDY-based and pin-related demand maps~\cite{spindler2007fast}; the supply group includes GR and eGR utilization or overflow-related signals from the routing flow~\cite{pan2006fastroute,liu2013nctu}; and the geometry group includes macro obstacles, cell density, and macro-boundary context. This grouping keeps the conditioning design simple, consistent, and physically grounded.

Table~\ref{tab:channels} lists the final task-specific control sets. The expanded control banks contain 11 channels for congestion and 16 channels for DRC. After removing one inactive channel per task before training, the model uses 10 effective congestion controls and 15 effective DRC controls. For congestion, the control set combines demand and supply signals so that the model can observe both where wires tend to accumulate and how much routing capacity remains available. For DRC, the control set retains demand, utilization, overflow, and geometry-related signals, since DRC violations depend on more localized interactions among these factors. In both tasks, the original CircuitNet controls are retained, and the added controls extend them with information that is more directly tied to routing resources and geometric constraints.

To support this design, we compute the Pearson correlation between each routing-control channel $f_c$ and the target label $y$ over all pixels,
\begin{equation}
r_c = \mathrm{corr}(f_c, y).
\label{eq:channel_corr}
\end{equation}
For congestion, the strongest global correlations appear in both supply and demand channels, including $\texttt{GR\_util\_H}$ ($r=0.4831$), $\texttt{RUDY\_pin}$ ($r=0.4358$), $\texttt{GR\_util\_V}$ ($r=0.4007$), and $\texttt{cell\_density}$ ($r=0.3885$). This supports using routing controls that expose both routing pressure and remaining routing resources. For DRC, no single channel dominates. Instead, the signal is distributed across several moderate-correlation channels, including $\texttt{GR\_overflow\_V}$ ($r=0.2865$), $\texttt{GR\_overflow\_H}$ ($r=0.2681$), $\texttt{eGR\_util\_V}$ ($r=0.2236$), and $\texttt{GR\_util\_H}$ ($r=0.1999$). This supports using a broader task-specific control set for DRC. Additional dataset analysis is used in the next section to motivate the latent design for each task.

All routing controls are rasterized on the same spatial grid and normalized to the same range before they are passed to the model. Figure~\ref{fig:drc_controls} shows representative examples from these control groups together with the target map. The main point of this subsection is that the routing controls are selected from physically meaningful signal groups and tailored to the different structures of congestion and DRC.

\begin{table}[t]
\centering
\caption{Task-specific routing controls used in the experiments. The reported correlation is the global Pearson correlation $r_c=\mathrm{corr}(f_c,y)$ on the N28 expanded dataset. Channels are grouped by physical role.}
\label{tab:channels}
\footnotesize
\setlength{\tabcolsep}{15pt}
\begin{tabular}{lllc}
\toprule
Task & Channel & Group & Corr. $r_c$ \\
\midrule
\multirow{10}{*}{Cong}
& \texttt{macro\_region}    & Geometry & -0.2768 \\
& \texttt{RUDY}             & Demand   &  0.3855 \\
& \texttt{RUDY\_pin}        & Demand   &  0.4358 \\
& \texttt{RUDY\_long}       & Demand   &  0.3754 \\
& \texttt{RUDY\_short}      & Demand   &  0.2577 \\
& \texttt{cell\_density}    & Geometry &  0.3885 \\
& \texttt{GR\_util\_H}      & Supply   &  0.4831 \\
& \texttt{GR\_util\_V}      & Supply   &  0.4007 \\
& \texttt{eGR\_overflow\_H} & Supply   &  0.0404 \\
& \texttt{eGR\_overflow\_V} & Supply   &  0.0900 \\
\midrule
\multirow{15}{*}{DRC}
& \texttt{macro\_region}    & Geometry & -0.1016 \\
& \texttt{cell\_density}    & Geometry &  0.1396 \\
& \texttt{RUDY\_long}       & Demand   &  0.1748 \\
& \texttt{RUDY\_short}      & Demand   &  0.0949 \\
& \texttt{RUDY\_pin\_long}  & Demand   &  0.1485 \\
& \texttt{eGR\_overflow\_H} & Supply   &  0.0407 \\
& \texttt{eGR\_overflow\_V} & Supply   &  0.0640 \\
& \texttt{GR\_overflow\_H}  & Supply   &  0.2681 \\
& \texttt{GR\_overflow\_V}  & Supply   &  0.2865 \\
& \texttt{GR\_util\_H}      & Supply   &  0.1999 \\
& \texttt{GR\_util\_V}      & Supply   &  0.1524 \\
& \texttt{RUDY}             & Demand   &  0.1774 \\
& \texttt{RUDY\_pin}        & Demand   &  0.1503 \\
& \texttt{eGR\_util\_H}     & Supply   &  0.2112 \\
& \texttt{eGR\_util\_V}     & Supply   &  0.2236 \\
\bottomrule
\end{tabular}
\end{table}

\subsection{Task-Specific Latent Encoding for Routability Fields}
\label{sec:latent_design}

\begin{figure}[t]
    \centering
    \includegraphics[width=\columnwidth]{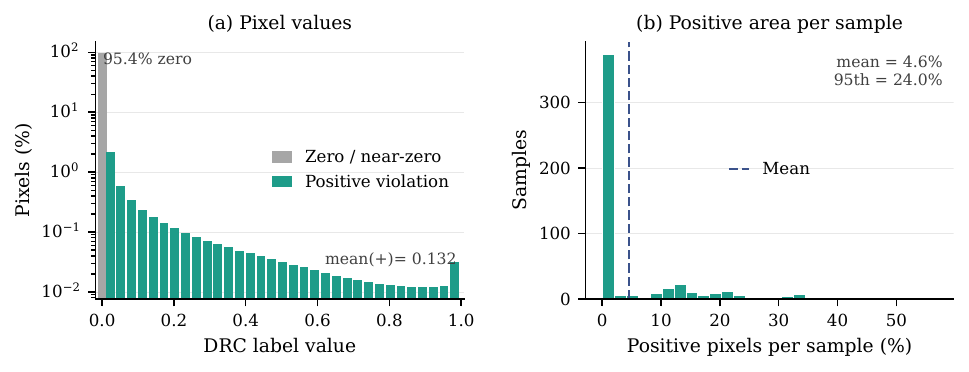}
    \caption{Distribution of DRC labels in the training set. The target is highly sparse: most pixels are zero or near-zero, and the positive violation area varies substantially across designs.}
    \label{fig:drc_dataset_characterization}
\end{figure}

\begin{figure}[t]
    \centering
    \includegraphics[width=\columnwidth]{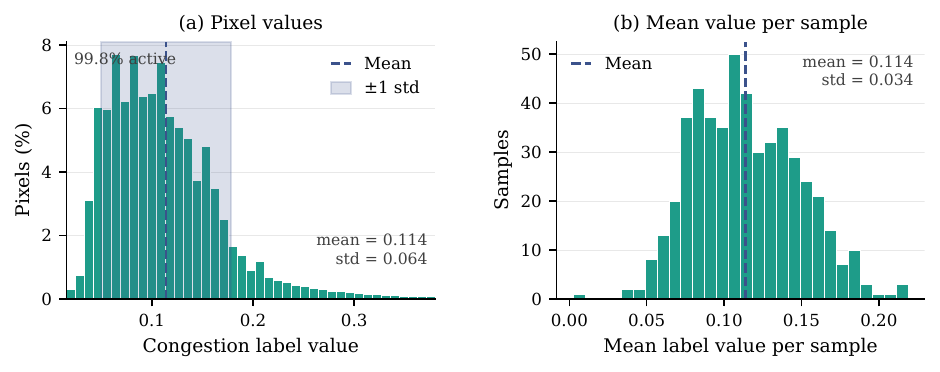}
    \caption{Distribution of congestion labels in the training set. The target is dense and concentrated around a relatively narrow value range, with moderate variation across designs.}
    \label{fig:cong_dataset_characterization}
\end{figure}

Raw pixel space is not an ideal domain for routability-field generation. A natural alternative is to first encode the target map into a compact latent representation with a variational autoencoder (VAE)~\cite{kingma2013auto}, and then perform diffusion in that latent space~\cite{Rombach2021HighResolutionIS}. In our setting, however, a single generic latent design is not well matched to both targets.

The main reason is that DRC and congestion have very different label structures. As shown in Fig.~\ref{fig:drc_dataset_characterization}(a), the DRC target is highly sparse: 95.4\% of pixels are zero, the mean positive value is 0.132, and the mean positive area per sample is only 4.6\% (Fig.~\ref{fig:drc_dataset_characterization}(b)). In contrast, Fig.~\ref{fig:cong_dataset_characterization}(a) shows that the congestion target is dense: 99.8\% of pixels are active, with values concentrated around a mean of 0.114 and standard deviation of 0.064 (Fig.~\ref{fig:cong_dataset_characterization}(b)).
These different target statistics lead to different latent-design requirements. For DRC, the main challenge is to preserve sparse violation structure and rare high-value regions. For congestion, the main challenge is to represent a dense spatial field faithfully while maintaining a stable latent space for generation. This motivates task-specific latent encoding for the two routability fields.
We therefore use \emph{task-specific} latent compression before diffusion. In both tasks, the encoder maps a $1\times256\times256$ routing map to a latent posterior at $64\times64$ resolution, and the decoder maps the latent back to a $1\times256\times256$ output map. The latent resolution is chosen from the data. A $4\times$ spatial reduction from $256\times256$ to $64\times64$ preserves essentially all task-relevant signal for both targets, whereas a more aggressive $8\times$ reduction to $32\times32$ begins to discard sparse DRC structure. The VAE is therefore not only a computational tool; it is the interface that converts each routing target into a generation domain better matched to its statistics.

Both encoders use the same high-level objective
\begin{equation}
  \mathcal{L}_{\mathrm{VAE}}^{(\mathrm{task})}
  \;=\;
  \mathcal{L}_{\mathrm{recon}}^{(\mathrm{task})}
  \;+\;
  \beta\,\mathcal{L}_{\mathrm{KL}},
  \label{eq:vae_objective}
\end{equation}
where the reconstruction term depends on the task and the KL term keeps the latent compatible with the Gaussian prior used by the diffusion model. We use per-channel free bits~\cite{kingma2013auto}:
\begin{equation}
  \mathcal{L}_{\mathrm{KL}}
  \;=\;
  \frac{1}{C_z}
  \sum_{c=1}^{C_z}
  \max\!\bigl(\lambda,\mathrm{KL}_c\bigr),
  \qquad
  \lambda = 0.5 \text{ nats},
  \label{eq:free_bits}
\end{equation}
where $C_z$ is the number of latent channels and $\mathrm{KL}_c$ is the KL divergence of channel $c$. This prevents individual channels from collapsing too early and keeps the latent usable for diffusion.

\subsubsection{Encoding Sparse DRC Violation Maps}
\label{sec:drc_vae}

The DRC violation map requires a violation-focused latent design. The difficulty is not only that nonzero violations are rare, but also that most of the map is zero background. A uniform reconstruction loss in raw pixel space would therefore spend most of its capacity on background pixels instead of the spatial violation pattern.

We address this in three steps. First, we encode the raw DRC violation map $x_{\mathrm{DRC}}\in[0,1]^{256\times256}$ in log space:
\begin{equation}
  \tilde{x}_{\mathrm{DRC}}
  \;=\;
  \log\!\left(1 + 10\,x_{\mathrm{DRC}}\right).
  \label{eq:drc_log}
\end{equation}
This expands the low-value nonzero region before compression and makes small violations easier to separate from the background.

Second, the decoder reconstructs the output map $\hat{x}_{\mathrm{DRC}}\in[0,1]^{256\times256}$ with a violation-weighted loss:
\begin{equation}
  \mathcal{L}_{\mathrm{focal}}^{\mathrm{DRC}}
  \;=\;
  \frac{1}{HW}
  \sum_{i=1}^{HW}
  \Bigl(1 + \gamma \sqrt{x_{\mathrm{DRC},i}}\Bigr)
  \bigl(\hat{x}_{\mathrm{DRC},i} - x_{\mathrm{DRC},i}\bigr)^2,
  \qquad
  \gamma = 20.
  \label{eq:drc_focal}
\end{equation}
This increases the training signal on nonzero violation pixels and reduces the dominance of the zero background.

Third, we add a hotspot term on the top $1\%$ highest-valued DRC pixels in each sample:
\begin{equation}
  \mathcal{L}_{\mathrm{hot}}^{\mathrm{DRC}}
  \;=\;
  \mathrm{MSE}\!\left(
  \hat{x}_{\mathrm{DRC}}^{\mathrm{top}\,1\%},
  x_{\mathrm{DRC}}^{\mathrm{top}\,1\%}
  \right),
  \label{eq:drc_hot}
\end{equation}
and use the full DRC reconstruction objective
\begin{equation}
  \mathcal{L}_{\mathrm{recon}}^{\mathrm{DRC}}
  \;=\;
  \mathcal{L}_{\mathrm{focal}}^{\mathrm{DRC}}
  \;+\;
  0.5\,\mathcal{L}_{\mathrm{hot}}^{\mathrm{DRC}}.
  \label{eq:drc_recon}
\end{equation}

The DRC encoder compresses the $1\times256\times256$ input map to a $12\times64\times64$ latent and uses $\beta=0.05$ with a 15-epoch warmup. The purpose of this design is to preserve rare violation structure while still producing a latent compatible with diffusion. Table~\ref{tab:latent_stats} shows that all 12 latent channels remain active, with posterior standard deviations in the narrow range $[0.630,\,0.642]$, indicating that the DRC latent stays well distributed instead of collapsing to a few channels.

\subsubsection{Encoding Dense Congestion Maps}
\label{sec:cong_vae}

The congestion map requires a different latent design. As shown in Fig.~\ref{fig:cong_dataset_characterization}, the congestion label is dense and smooth, not sparse. The main failure mode is therefore not background dominance, but posterior collapse: if the KL pressure is too strong, the encoder can represent the dense congestion field with only a few dominant channels and let the remaining channels match the prior~\cite{lucas2019don,ichikawa2024learning}.

For this reason, we keep the reconstruction term simple. Let $x_{\mathrm{cong}}\in[0,1]^{256\times256}$ be the raw congestion map and $\hat{x}_{\mathrm{cong}}\in[0,1]^{256\times256}$ its reconstruction. We use
\begin{equation}
  \mathcal{L}_{\mathrm{recon}}^{\mathrm{cong}}
  \;=\;
  \frac{1}{HW}
  \sum_{i=1}^{HW}
  \left|\hat{x}_{\mathrm{cong},i} - x_{\mathrm{cong},i}\right|.
  \label{eq:cong_l1}
\end{equation}
No log transform, focal weighting, or hotspot term is used for congestion.

The main design effort is placed on the latent regularization. The congestion encoder compresses the $1\times256\times256$ input map to an $8\times64\times64$ latent. We again use the free-bits KL term in Eq.~\eqref{eq:free_bits}, but with a smaller KL weight, $\beta=0.005$, and a longer 40-epoch warmup. This reduces the pressure to compress the dense field into only a few latent dimensions.

Table~\ref{tab:cong_vae} shows why this design is necessary. With a stronger KL setting, only $3$ of $8$ channels remain active. With the final design, all $8$ channels remain active, with posterior standard deviations in the narrow range $[0.784,\,0.801]$. For congestion, the VAE therefore solves a different problem than for DRC: not rare-event preservation, but stable dense-field compression without latent collapse.

\begin{table}[t]
\centering
\caption{Effect of KL design on congestion latent quality. Active channels have $\sigma_c > 0.10$. The final design keeps all 8 channels active and produces a latent suitable for diffusion.}
\label{tab:cong_vae}
\footnotesize
\setlength{\tabcolsep}{5pt}
\begin{tabular}{lcccc}
\toprule
Configuration & $\beta$ & Free-bits ($\lambda$) &
Active channels & $\sigma_c$ range \\
\midrule
Standard VAE  & $0.020$ & None       & $3/8$ & $[0.008,\;0.480]$ \\
Ours          & $0.005$ & $0.5$ nats & $\mathbf{8/8}$ &
$[0.784,\;0.801]$ \\
\bottomrule
\end{tabular}
\end{table}



Table~\ref{tab:latent_stats} reports the latent statistics before normalization. In both tasks, the channel means are already close to zero and the channel standard deviations lie in a narrow range. After normalization, the conditional diffusion model receives a standardized latent target for both DRC and congestion.
The final latent targets for diffusion are therefore $12\times64\times64$ for DRC violation maps and $8\times64\times64$ for congestion maps. These latents preserve the task-relevant spatial signal while moving generation away from the raw pixel domain.
\begin{table}[t]
\centering
\caption{Latent channel statistics before normalization for both task-specific encoders. Near-zero channel means and narrow standard-deviation ranges indicate that both latents are well behaved before handoff to the conditional diffusion model.}
\label{tab:latent_stats}
\footnotesize
\setlength{\tabcolsep}{5pt}
\begin{tabular}{llcccc}
\toprule
Encoder & Task & $C_z$ &
$\sigma_c$ range & $\max|\mu_c|$ & Global $\bar{\sigma}$ \\
\midrule
DRC encoder        & DRC        & $12$ &
$[0.630,\;0.642]$ & $0.008$ & $0.637$ \\
Congestion encoder & Congestion & $8$  &
$[0.784,\;0.801]$ & $0.017$ & $0.792$ \\
\bottomrule
\end{tabular}
\end{table}

\subsection{Conditional Latent Diffusion with Multi-Scale Routing Controls}
\label{sec:ldm}

After learning a task-specific latent space for each target map, we modeled routability-map generation with a conditional latent diffusion model. Latent diffusion is a standard framework for generative modeling in compressed latent spaces~\cite{Rombach2021HighResolutionIS,Ho2020DenoisingDP,salimans2022progressive}. Our focus here was the conditioning design for routability generation. Instead of treating placement-stage routing features as a single appended input, we used them as explicit \emph{routing controls} that guided generation at multiple spatial scales.

Let $x \in \mathbb{R}^{1\times 256\times 256}$ denote the target routability map and let $f \in \mathbb{R}^{C_{\mathrm{feat}}\times 256\times 256}$ denote the routing feature tensor. The task-specific VAE encoder mapped $x$ to a compact latent code $z_0 \in \mathbb{R}^{C_z\times 64\times 64}$, and the diffusion model learned the conditional distribution $p(z_0 \mid f)$. The denoising UNet took only the noisy latent as input. The routing features were processed by a separate control branch and injected into the UNet internally. This design made the conditioning role explicit: the latent represented \emph{what} should be generated, while the routing controls determined \emph{how} that generation should follow the physical structure of the design.

A key observation was that routing pressure is naturally multi-scale. Global demand affects the coarse layout of congestion and violation regions, intermediate demand shapes regional structure, and local overflow-related signals influence the precise placement of hotspots. For this reason, we did not use a single conditioning tensor. Instead, as shown in Fig.~\ref{fig:overview}(c), we transformed the routing features into three control tensors, \(c_{64}\), \(c_{32}\), and \(c_{16}\), and injected them into the corresponding stages of the latent UNet.

\noindent\textbf{Multi-scale routing controls.}
A lightweight feature encoder transforms the routing feature map into three spatial resolutions,
\begin{equation}
\left(f_{64},\, f_{32},\, f_{16}\right) = \mathcal{C}(f),
\label{eq:multiscale_feat}
\end{equation}
where the subscripts denote spatial size. Each feature map is then projected to the channel dimension required by the corresponding UNet stage using a zero-initialized \(1\times1\) convolution,
\begin{equation}
c_s = W_s * f_s, \qquad s \in \{64,32,16\},
\label{eq:zeroconv_cond}
\end{equation}
with \(W_s\) initialized to zero. The resulting routing controls are injected at matched resolutions: \(c_{64}\) is added to the high-resolution skip connection, \(c_{32}\) is added to the intermediate skip connection, and \(c_{16}\) is injected at the bottleneck before the middle attention block. This lets the model use coarse routing context to shape global structure and finer routing context to refine local map details.

The zero-initialized projections make the control branch stable to train. At initialization, all routing controls are exactly zero, so the network first behaves as a latent denoiser without feature guidance. As training proceeds, the control branch gradually learns how to modulate generation using routing signals. We also use classifier-free guidance by dropping all three control tensors together with probability \(p_{\mathrm{cfg}}\). Each training sample is therefore seen either with full routing control or with no routing control, which keeps the conditional and unconditional branches consistent across scales.

Given the normalized latent target \(\hat{z}_0\), a timestep \(t\), and Gaussian noise \(\epsilon\), we form the noisy latent \(\hat{z}_t\) using the standard forward diffusion process. The denoising network \(v_{\theta}\) then predicts the velocity target in latent space conditioned on the routing controls:
\begin{equation}
\mathcal{L}_{\mathrm{ldm}}
=
\mathbb{E}_{t,\hat{z}_0,\epsilon}
\left[
\left\|
v_{\theta}\!\left(\hat{z}_t, t; c_{64}, c_{32}, c_{16}\right) - v
\right\|^2
\right],
\label{eq:ldm_loss_control}
\end{equation}
where
\begin{equation}
v = \sqrt{\bar{\alpha}_t}\,\epsilon - \sqrt{1-\bar{\alpha}_t}\,\hat{z}_0 .
\label{eq:velocity_target_control}
\end{equation}
This objective follows the standard latent diffusion formulation~\cite{Rombach2021HighResolutionIS,Ho2020DenoisingDP,salimans2022progressive}. The task-specific component in our setting is the use of multi-scale routing controls, which guide denoising toward routability maps that remain consistent with placement-stage signals. Algorithm~\ref{alg:cldroute_training} summarizes the training and inference procedure of the task-specific routing-controlled latent diffusion model.

\begin{algorithm}[t]
\caption{Task-specific routing-controlled latent diffusion}
\label{alg:cldroute_training}
\small
\begin{algorithmic}[1]
\Require routing controls $f$, target routability map $x$, task-specific VAE encoder $\mathrm{Enc}$, task-specific VAE decoder $\mathrm{Dec}$, multi-scale conditioner $\mathcal{C}$, denoiser $v_\theta$
\State $z_0 \leftarrow \mathrm{Enc}(x)$
\State sample timestep $t$ and Gaussian noise $\epsilon$
\State construct noisy latent $z_t$ from $z_0$, $t$, and $\epsilon$
\State $(f_{64}, f_{32}, f_{16}) \leftarrow \mathcal{C}(f)$
\State $(c_{64}, c_{32}, c_{16}) \leftarrow (W_{64} * f_{64},\, W_{32} * f_{32},\, W_{16} * f_{16})$
\State apply joint classifier-free guidance dropout to $(c_{64}, c_{32}, c_{16})$
\State $\hat{v} \leftarrow v_{\theta}(z_t, t; c_{64}, c_{32}, c_{16})$
\State update model parameters using $\|\hat{v} - v\|^2$
\Statex
\State \textbf{Inference:} for a fixed routing-control tensor $f$, generate $N$ latent samples $\{z^{(k)}\}_{k=1}^{N}$
\State decode each sample $x^{(k)} \leftarrow \mathrm{Dec}(z^{(k)})$
\State return mean map $\bar{x}=\frac{1}{N}\sum_{k=1}^{N} x^{(k)}$ and variance map $\sigma^2=\mathrm{Var}_k[x^{(k)}]$
\end{algorithmic}
\end{algorithm}

\noindent\textbf{Sample-based generation and uncertainty.}
At inference time, we generate multiple latent samples for the same routing-control tensor \(f\) and decode them back to label space with the task-specific VAE decoder. The final routability estimate is the sample mean
\begin{equation}
\bar{x}
=
\frac{1}{N}\sum_{k=1}^{N}\mathrm{Dec}\!\left(z^{(k)}\right),
\label{eq:ldm_mean}
\end{equation}
and the accompanying uncertainty map is the per-pixel sample variance
This gives two outputs for the same placed design: a mean routability field and a spatial map showing where the generated outcome varies more across samples. From the application point of view, the model therefore provides both a likely routability map and a direct indication of where the placement-stage signals support multiple plausible outcomes.
\section{Evaluation}

\noindent\textbf{Experimental Setup.}
All experiments were performed on a Linux system with an AMD EPYC 7763
64-core processor and four NVIDIA RTX A6000 GPUs (48\,GB each).
The task specific VAEs train with AdamW,
learning rate $10^{-4}$, cosine decay with 500-step linear warmup, EMA decay 0.999.
The latent diffusion model with multi-scale ControlNet conditioning
trains with the same optimizer and EMA settings
$T=1000$ diffusion steps, cosine noise schedule~\cite{Ho2020DenoisingDP}.
At inference, DDIM~\cite{song2020denoising} with $N=8$
independent draws per design, averaged in label space after VAE
decoding.

\noindent\textbf{Dataset.}
We evaluate on the CircuitNet~2.0 dataset~\cite{jiang2024circuitnet,chai2023circuitnet} at two technology nodes, N28 and N14, corresponding to 28\,nm and 14\,nm designs. We use a design-wise split for both congestion and DRC, so that training, validation, and test sets contain disjoint designs. The split sizes are summarized in Table~\ref{tab:data_split}. 
\begin{table}[t]
\centering
\caption{Design-wise dataset splits used in the experiments.}
\label{tab:data_split}
\footnotesize
\setlength{\tabcolsep}{20pt}
\begin{tabular}{lccc}
\toprule
Node & Train & Val & Test \\
\midrule
N28 & 7,872  & 1,248 & 1,122 \\
N14 & 10,368 & 169   & 250 \\
\bottomrule
\end{tabular}
\end{table}

\subsection{Evaluation Metrics}
We reported two groups of metrics. The first group consisted of standard regression metrics used for cross-method comparison. The second group consisted of task-specific metrics designed to better reflect how the generated routability fields would be used in physical design.

\noindent\textbf{Standard metrics.}
We report MAE, NRMS, SSIM, and Pearson correlation. MAE and NRMS measured pixel-wise reconstruction error, SSIM measured structural similarity between the generated and target maps, and Pearson correlation measured linear agreement over the full field. These metrics provided a common basis for comparison with prior work, but they did not fully capture sign-off-oriented behavior, especially when the target map was sparse.

\noindent\textbf{Task-specific metrics.}
We additionally reported metrics that were aligned with the physical meaning of each task. For DRC, the main question was whether the model identified the correct violation regions and estimated their severity well. For congestion, the main question was whether the generated field preserved the spatial distribution of routing demand and remained well aligned in active regions.
For DRC, we reported \textit{TopK@1\%} and \textit{TopK@0.5\%}, which measured the overlap between the highest-valued predicted locations and the highest-valued ground-truth locations. These metrics evaluated hotspot localization directly. We also reported \textit{F1@0.1}, which summarized precision and recall at a fixed severity threshold, \textit{Hotspot-MAE}, which measured error restricted to ground-truth hotspot regions, and \textit{NZ-Pearson}, which computed correlation only on nonzero ground-truth pixels to reduce the effect of the dominant background.
For congestion, we reported \textit{NZ-Pearson}, \textit{Spatial Bias}, and 
\textit{Uncertainty-error correlation}. NZ-Pearson measured correlation in active regions of the congestion field. Spatial Bias measured the signed difference between the predicted and target global congestion levels, where values closer to zero indicated better overall calibration of the generated field. Uncertainty-error correlation, reported for generative models, measured the Pearson correlation between per-pixel predictive variance across multiple samples and the corresponding absolute error. Higher values indicated that regions with larger predictive uncertainty also tended to have larger generation error, making the uncertainty map more informative in practice.
Overall, the task-specific metrics complemented the standard metrics by evaluating aspects of routability quality that were more directly related to downstream physical-design use.
\subsection{Variational Autoencoder Evaluation}
\label{sec:vae_results}

We evaluate the task-specific VAEs on held-out test data to verify that each model preserves the structure of its target label space before diffusion. Table~\ref{tab:vae_results} summarizes the reconstruction results on N28 and N14.
On N28, both task-specific VAEs show high reconstruction fidelity. For DRC, the VAE reaches an MAE of 0.00095, an SSIM of 0.9934, and a correlation of 0.9870. For congestion, it reaches an MAE of 0.00132, an SSIM of 0.9932, and a correlation of 0.9608. These results indicate that the first-stage latent representations retain the main spatial structure of both routability fields on N28.
On N14, the congestion VAE also preserves the dense spatial structure well, with an MAE of 0.00390, an SSIM of 0.9676, and a Pearson correlation of 0.9305. For DRC, the target maps remain much sparser, so the reported correlation is NZ-Pearson. In this setting, the DRC VAE reaches an MAE of 0.00648, an SSIM of 0.7009, and an NZ-Pearson of 0.2782. Together, these results show that the task-specific first stage adapts to both dense congestion fields and sparse DRC maps across technology nodes.
\begin{table}[t]
\centering
\caption{Reconstruction quality of the task-specific VAEs}
\label{tab:vae_results}
\footnotesize
\setlength{\tabcolsep}{8pt}
\begin{tabular}{llccc}
\toprule
Dataset & Task & MAE $\downarrow$ & SSIM $\uparrow$ & Correlation $\uparrow$ \\
\midrule
\multirow{2}{*}{N28}
  & DRC        & 0.00095 & 0.9934 & 0.9870 \\
  & Congestion & 0.00132 & 0.9932 & 0.9608 \\
\midrule
\multirow{2}{*}{N14}
  & DRC        & 0.00648 & 0.7009 & 0.2782 \\
  & Congestion & 0.00390 & 0.9676 & 0.9305 \\
\bottomrule
\end{tabular}
\end{table}

\section{Routability Evaluation}
\label{sec:results_n28}

We report both standard metrics and task-specific metrics. The standard metrics maintain direct comparability with prior routability literature, while the task-specific metrics reflect how the generated fields are used in physical design. For DRC, the key questions are whether the method localizes violation hotspots and estimates their severity accurately. For congestion, the key questions are whether the generated field captures the spatial distribution of routing demand and remains well aligned in active regions.

\subsection{DRC violation map generation on N28}
\begin{table}[t]
\centering
\caption{Standard metrics on the N28 DRC task.}
\label{tab:drc_n28_standard}
\resizebox{\columnwidth}{!}{%
\begin{tabular}{lcccc}
\toprule
Method & MAE $\downarrow$ & NRMS $\downarrow$ & SSIM $\uparrow$ & Pearson $\uparrow$ \\
\midrule
RouteNet (SOTA) & 0.00313 & 0.03300 & 0.95481 & 0.38156 \\
Pixel Diffusion (9ch) & 0.01961 & 0.20180 & 0.59270 & 0.28990 \\
LDM (15ch) & 0.00292 & 0.03373 & 0.96470 & 0.50477 \\
LDM+ControlNet (15ch) & \textbf{0.00280} & \textbf{0.02893} & \textbf{0.96780} & \textbf{0.52483} \\
\bottomrule
\end{tabular}%
}
\end{table}
Tables~\ref{tab:drc_n28_standard} and~\ref{tab:drc_n28_task} summarize the DRC results on CircuitNet~2.0 N28~\cite{jiang2024circuitnet}, with RouteNet~\cite{8587655} as the deterministic baseline. On the standard metrics, the latent generative models give the strongest overall results, and LDM+ControlNet achieves the best values on all four metrics: MAE \(=0.00280\), NRMS \(=0.02893\), SSIM \(=0.96780\), and Pearson \(=0.52483\). These values indicate that the generated DRC fields preserve both pixel-level fidelity and global spatial structure.

The task-specific metrics give the clearest view of DRC quality. LDM+ControlNet achieves the best TopK@1\% and TopK@0.5\%, with values of 0.34940 and 0.34830, showing precise localization of the highest-risk regions. It also gives the best Hotspot-MAE, 0.05580, and the best F1@0.1, 0.44203, indicating accurate support and severity modeling in active violation regions. NZ-Pearson remains high at 0.46143, which shows that the ranking of nonzero violation regions is also preserved well. Overall, the DRC results highlight the value of task-specific evaluation, since the hotspot-oriented metrics align closely with how DRC maps are used in sign-off analysis.

\begin{table}[t]
\centering
\caption{DRC-specific metrics on N28.}
\label{tab:drc_n28_task}
\resizebox{\columnwidth}{!}{%
\begin{tabular}{lccccc}
\toprule
Method & TopK@1\% $\uparrow$ & TopK@0.5\% $\uparrow$ & Hotspot-MAE $\downarrow$ & NZ-Pearson $\uparrow$ & F1@0.1 $\uparrow$ \\
\midrule
RouteNet & 0.24164 & 0.25454 & 0.05729 & 0.39655 & 0.41343 \\
Pixel Diffusion (9ch) & 0.22073 & 0.24693 & 0.07683 & \textbf{0.46287} & 0.15370 \\
LDM (15ch) & 0.33837 & 0.33250 & 0.06013 & 0.44733 & 0.40247 \\
LDM+Cnt (15ch) & \textbf{0.34940} & \textbf{0.34830} & \textbf{0.05580} & 0.46143 & \textbf{0.44203} \\
\bottomrule
\end{tabular}%
}
\end{table}

\subsection{Congestion map generation on N28}

Tables~\ref{tab:cong_n28_standard} and~\ref{tab:cong_n28_task} summarize the congestion results on CircuitNet~2.0 N28~\cite{jiang2024circuitnet}, with GPDL~\cite{9473959} as the prior baseline. On the standard dense-field metrics, Pixel Diffusion achieves the best MAE, NRMS, and SSIM, with values of 0.02730, 0.03167, and 0.91990, respectively. LDM+ControlNet achieves the best Pearson correlation, 0.36870, indicating the strongest global spatial agreement among the compared methods.
The task-specific congestion metrics further clarify the behavior of the generative models. LDM+ControlNet achieves the best NZ-Pearson, 0.36923, the spatial bias closest to zero, \(-0.00441\), and the highest uncertainty score, 0.35920. These values show that its generated congestion fields align well with the active spatial demand pattern, remain close to an unbiased estimate overall, and provide informative uncertainty. Together, the congestion results show that the standard metrics capture dense-field reconstruction quality, while the task-specific metrics capture spatial demand fidelity and uncertainty quality more directly.

\begin{table}[t]
\centering
\caption{Standard regression metrics on the N28 congestion task. Best results are shown in bold.}
\label{tab:cong_n28_standard}
\resizebox{\columnwidth}{!}{%
\begin{tabular}{lcccc}
\toprule
Method & MAE $\downarrow$ & NRMS $\downarrow$ & SSIM $\uparrow$ & Pearson $\uparrow$ \\
\midrule
GPDL (SOTA) & 0.02958 & 0.03367 & 0.90960 & 0.25433 \\
Pixel Diffusion (3ch) & \textbf{0.02730} & \textbf{0.03167} & \textbf{0.91990} & 0.30770 \\
LDM (10ch) & 0.02915 & 0.03380 & 0.91223 & 0.33127 \\
LDM+ControlNet (10ch) & 0.02859 & 0.03430 & 0.90310 & \textbf{0.36870} \\
\bottomrule
\end{tabular}%
}
\end{table}

\begin{table}[t]
\centering
\caption{Congestion metrics on N28. For Spatial Bias, values closer to zero are better.}
\label{tab:cong_n28_task}
\resizebox{\columnwidth}{!}{%
\begin{tabular}{lccc}
\toprule
Method & NZ-Pearson $\uparrow$ & Spatial Bias $\rightarrow 0$ & Uncertainty $\uparrow$ \\
\midrule
GPDL (SOTA) & 0.25487 & 0.00667 & -- \\
Pixel Diffusion (3ch) & 0.30883 & 0.00708 & 0.24107 \\
LDM (10ch) & 0.33167 & -0.01259 & 0.21233 \\
LDM+Cnt (10ch) & \textbf{0.36923} & \textbf{-0.00441} & \textbf{0.35920} \\
\bottomrule
\end{tabular}%
}
\end{table}

\subsection{DRC violation map generation on N14}
\begin{table}[t]
\centering
\caption{N14 DRC generation results.}
\label{tab:n14_drc_results}
\resizebox{\columnwidth}{!}{%
\begin{tabular}{lcccccc}
\toprule
Method & MAE $\downarrow$ & SSIM $\uparrow$ & TopK@1\% $\uparrow$ & Hotspot-MAE $\downarrow$ & NZ-Pearson $\uparrow$ & Uncertainty $\uparrow$ \\
\midrule
LDM  & \textbf{0.00627} & \textbf{0.7136} & 0.0125 & \textbf{0.00571} & \textbf{0.0874} & \textbf{0.4972} \\
LDM+Ctn  & 0.00633 & 0.7089 & \textbf{0.0148} & 0.00578 & 0.0358 & 0.4329 \\
\bottomrule
\end{tabular}%
}
\end{table}

Table~\ref{tab:n14_drc_results} shows that the framework transfers to N14 for DRC generation. The unified LDM gives the strongest overall result profile, with the best MAE (0.00627), SSIM (0.7136), Hotspot-MAE (0.00571), NZ-Pearson (0.0874), and uncertainty-error correlation (0.4972). The ControlNet variant gives the best TopK@1\% (0.0148), indicating slightly sharper localization of the highest-risk regions. Since the N14 DRC labels are sparse and have a limited dynamic range, the hotspot-oriented metrics are the more informative indicators here. Overall, the N14 results show that the same conditional generation framework remains effective across technology nodes, with the unified latent model giving the more balanced DRC behavior.

\subsection{Congestion map generation on N14}
\begin{table}[t]
\centering
\caption{N14 congestion generation results.}
\label{tab:n14_cong_results}
\resizebox{\columnwidth}{!}{%
\begin{tabular}{lccccc}
\toprule
Method & MAE $\downarrow$ & SSIM $\uparrow$ & Pearson $\uparrow$ & Spatial Bias $\rightarrow 0$ & Uncertainty $\uparrow$ \\
\midrule
LDM  & \textbf{0.03416} & \textbf{0.7678} & 0.0369 & -0.00472 & 0.0094 \\
LDM+ControlNet  & 0.03588 & 0.7654 & \textbf{0.0370} & \textbf{0.00297} & \textbf{0.0197} \\
\bottomrule
\end{tabular}%
}
\end{table}

Table~\ref{tab:n14_cong_results} shows stable transfer to N14 for congestion generation. The unified LDM gives the best dense-field reconstruction quality, with the lowest MAE (0.03416) and highest SSIM (0.7678). The ControlNet variant gives the best Pearson correlation (0.0370), the spatial bias closest to zero (0.00297), and the highest uncertainty-error correlation (0.0197). These numbers separate the roles of the two variants clearly: the unified model gives slightly better reconstruction fidelity, while the ControlNet model gives slightly better global alignment and uncertainty behavior. Taken together, the N14 results show that the same conditional generative framework remains usable across both N28 and N14.

Across both tasks, the N28 evaluation highlights two complementary aspects of routability-map generation. The first is standard fidelity, which enables direct comparison under a shared metric set. The second is task-aligned utility, which measures whether the generated maps emphasize the spatial regions that matter most for routing analysis. This distinction is especially important for DRC, where sparse violation structure makes hotspot-focused evaluation essential, and it remains equally informative for congestion, where the accuracy of high-demand regions and systematic demand balance strongly influence practical usefulness. The two-table evaluation protocol therefore provides a consistent and physically meaningful view of generation quality across both tasks.

\section{Conclusion}
This paper presents \emph{CLDRoute}, a conditional latent diffusion framework for routability map generation during physical design. Instead of treating routability estimation as deterministic prediction, CLDRoute models both congestion and DRC as spatially structured routability fields and generates them conditionally from placement-stage routing controls. The method combines physics-aware routing controls, task-specific latent encoding, and multi-scale conditional diffusion to handle the different structures of dense congestion maps and sparse DRC violation maps within one unified framework. Across the experiments, the results show that this formulation supports accurate generation on CircuitNet~2.0 N28 and also transfers to N14, while providing uncertainty estimates through sample-based inference. 
Overall, CLDRoute provides a practical placement-stage view of routability by generating both an expected routability map and a spatial uncertainty map from the same design representation.

 \bibliographystyle{ACM-Reference-Format}
  \bibliography{ref}
\end{document}